\documentclass[runningheads]{llncs}
\usepackage[T1]{fontenc}

\usepackage{url}
\usepackage{amsmath, amsfonts, amssymb, booktabs}
\usepackage{algorithm}
\usepackage{graphicx}
\usepackage{breqn}
\usepackage{listings}
\usepackage{algpseudocode}
\usepackage{multirow}
\usepackage{footnote}

\usepackage{xpatch}

\makeatletter
\xpatchcmd{\algorithmic}{\itemsep\z@}{\itemsep=0.3ex plus0pt}{}{}
\makeatother

\begin{document}
\mainmatter              
\title{BGM-HAN: A Hierarchical Attention Network for Accurate and Fair Decision Assessment on Semi-Structured Profiles}

\titlerunning{Byte-Pair Encoded, Gated Multi-head Hierarchical Attention Network}  
%
\author{Junhua Liu\inst{1,2}\orcidID{0000-0003-4477-7439}
\and Roy Ka-Wei Lee\inst{2}\orcidID{0000-0002-1986-7750}
\and Kwan Hui Lim\inst{2}\orcidID{0000-0002-4569-0901}}

\authorrunning{J. Liu et al.}  

\institute{
Singapore University of Technology and Design\\
\and
Forth AI\\
\email{j@forth.ai, roy\_lee@sutd.edu.sg, kwanhui\_lim@sutd.edu.sg}}

\maketitle 

Human decision-making in high-stakes domains often relies on expertise and heuristics, but is vulnerable to hard-to-detect cognitive biases that threaten fairness and long-term outcomes. This work presents a novel approach to enhancing complex decision-making workflows through the integration of hierarchical learning alongside various enhancements. Focusing on university admissions as a representative high-stakes domain, we propose BGM-HAN, an enhanced Byte-Pair Encoded, Gated Multi-head Hierarchical Attention Network, designed to effectively model semi-structured applicant data. BGM-HAN captures multi-level representations that are crucial for nuanced assessment, improving both interpretability and predictive performance. Experimental results on real admissions data demonstrate that our proposed model significantly outperforms both state-of-the-art baselines from traditional machine learning to large language models, offering a promising framework for augmenting decision-making in domains where structure, context, and fairness matter. Source code is available at: \url{https://github.com/junhua/bgm-han}.
\begin{figure}[t]
    \centering
    \includegraphics[width=.99\linewidth]{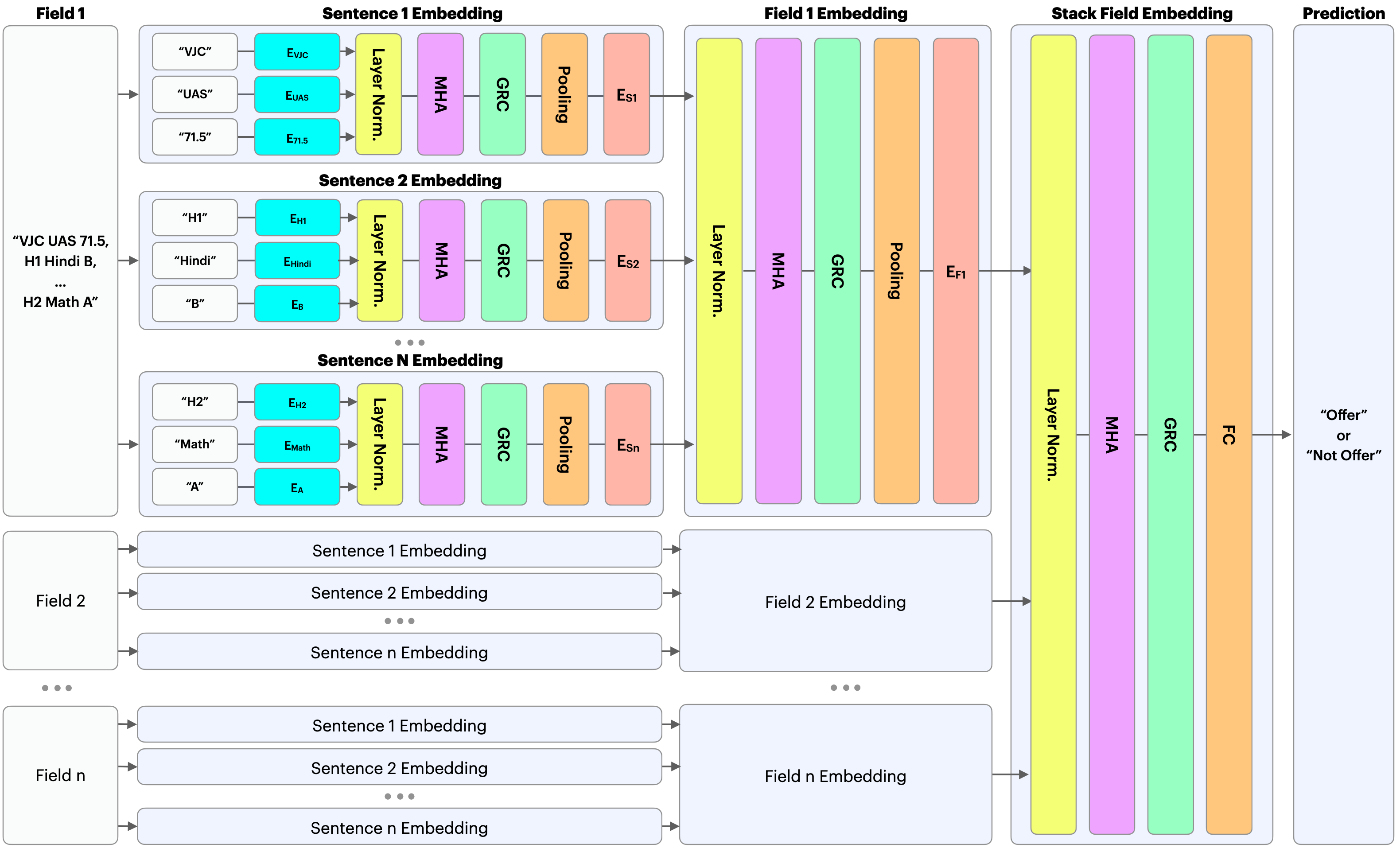}
    \caption{Architecture of the proposed BGM-HAN model. The multi-level model learns features from token to sentence to field. At each level, the data will go through layer normalisation, multi-head self-attention (MHA), gated residual connection (GRC), mean pooling to form the higher level embeddings. The embeddings are then concatenated and reshaped into 3D tensors to continue with the next level processing. }
    \label{fig:bgmhan}
\end{figure}

\section{Introduction}

High-stakes decision-making is often entrusted to human experts who rely on their domain knowledge and experiential judgment~\cite{alur2023auditing}. However, such decisions are susceptible to cognitive and affective biases, including anchoring~\cite{haag2024overcoming} and confirmation bias~\cite{echterhoff2024cognitive}, which are difficult to detect and mitigate~\cite{kahneman2023cognitive}. Addressing these biases is essential for ensuring fairness, transparency, and long-term sustainability, particularly in socially consequential domains~\cite{ghai2022d-bias}.

To mitigate human biases, recent research has explored the integration of artificial intelligence (AI) into human decision-making workflows. Notable approaches include fairness-aware AI systems that guide users toward more equitable decisions~\cite{yang2024fair}, explainable AI techniques that surface potential reasoning flaws~\cite{haag2024overcoming}, and human-AI collaborative frameworks for auditing social biases~\cite{ghai2022d-bias}.

Despite these advances, the practical impact of such systems remains limited due to several persistent challenges. First, the inherently context-dependent and latent nature of cognitive biases complicates their detection and correction by automated systems~\cite{kahneman2023cognitive}. Second, the limited interpretability of many AI models undermines user trust, which is an especially critical issue in high-stakes settings~\cite{haag2024overcoming}. Third, the scarcity of publicly available, domain-specific, and bias-sensitive datasets, which are often due to privacy or proprietary constraints thus posing substantial barriers to empirical progress~\cite{springer2024biasfairness}.

In this work, we address these limitations via the following contributions:

\begin{enumerate}
\item We introduce \textbf{BGM-HAN}, a model that incorporates {\textbf B}yte-Pair Encoding, {\textbf G}ated Residual Connections, and {\textbf M}ulti-Head Attention onto a {\textbf H}ierarchical {\textbf A}ttention {\textbf N}etwork. BGM-HAN is designed to effectively model semi-structured, multi-level data representations while maintaining interpretability.
\item We perform comprehensive empirical evaluations using a real-world university admissions dataset.\footnote{This is a proprietary dataset that is unable to be shared due to privacy concerns. The source codes however are publicly available at \url{https://github.com/junhua/bgm-han}.} Our analysis benchmarks BGM-HAN against state-of-the-art models, ranging from traditional machine learning models to neural networks and large language models. 
\item Results demonstrate that BGM-HAN significantly outperforms all baselines in terms of precision, recall, F1-score and accuracy, including state-of-the-art LLMs and human evaluators, thereby underscoring its potential to enhance decision quality in high-stakes applications.
\end{enumerate}

The remainder of this paper is structured as follows. Section~\ref{probFormulation} formally defines the problem of admission assessment. Section~\ref{proposedModel} presents the architecture and components of the proposed BGM-HAN model. Section~\ref{experimentMethodology} details the experimental methodology, including the setup, dataset, baseline models, and evaluation results. Section~\ref{relatedWork} reviews related literature and situates our work within the broader research landscape. Finally, Section~\ref{conclusion} concludes the paper and outlines directions for future research.

\section{Problem Formulation}
\label{probFormulation}

In high-stakes domains such as university admissions, decision-making involves evaluating complex, multifaceted student profiles under conditions of uncertainty and potential bias. Let the input space consist of a set of applicant profiles $\mathcal{P} = {p_1, \ldots, p_n}$, where each profile $p_i$ is composed of four principal components: (i) GCE A-Level results ($f_{\text{GCEA}}$), (ii) GCE O-Level results ($f_{\text{GCEO}}$), (iii) leadership records ($f_{\text{Leadership}}$), and (iv) responses to Personal Insight Questions (PIQs; $f_{\text{PIQ}}$). Each modality presents distinct representational challenges.

The academic records ($f_{\text{GCEA}}$, $f_{\text{GCEO}}$) contain structured grade data, often varying across subjects and examination cohorts, necessitating normalization and domain-aware scaling. The leadership records ($f_{\text{Leadership}}$) are semi-structured, comprising descriptive elements such as role titles, participation years, activity categories, and commitment levels. In contrast, the PIQ responses ($f_{\text{PIQ}}$) consist of unstructured free-form text aimed at assessing applicants' motivation, resilience, creativity, and alignment with institutional values, thereby demanding advanced natural language understanding capabilities.

The main objective is to learn a mapping function $\mathcal{D}: \mathcal{P} \rightarrow {0, 1}$ that maps each student profile to a binary admission outcome, where 1 denotes an offer and 0 a rejection. Crucially, the learned function must satisfy multiple real-world constraints: (i) \textit{fairness}: mitigating cognitive and algorithmic biases; (ii) \textit{consistency}: ensuring similar profiles yield similar decisions; and (iii) \textit{interpretability}: providing human-understandable rationales to support transparency and accountability in admissions.

\section{Proposed BGM-HAN Model}
\label{proposedModel}
The architecture overview of BGM-HAN is shown in Figure~\ref{fig:bgmhan}. We discuss the motivation and implementation of each key component in detail in the rest of this section. 

\subsection{Base Architecture}
BGM-HAN is designed on a base architecture inspired by Hierarchical Attention Network (HAN)~\cite{yang2016han}, which demonstrated proficiency in capturing the latent information of textual data where its structure embeds additional insights. 
This architecture aligns naturally with the semi-structured nature of university applicant profiles, which are composed of multi-level fields. For instance, academic records consist of multiple subject-grade pairs, and leadership experience comprises structured entries with attributes such as role, year, category, and participation level (refer to Section \ref{sectDataset} for more details).
HAN’s dual-level attention mechanisms at both entry and field levels enable the model to focus on the most informative parts of the text across hierarchy. We observe high relevance between neural architecture of HAN and the multi-level semi-structured nature of our data. This capability is particularly crucial for candidates assessments and decision recommendations, where key insights that influence decisions may be dispersed throughout different sections of an applicant’s profile. 

To enhance the base HAN, we integrate three key mechanisms: byte-pair encoding (BPE) for robust tokenization, multi-head self-attention~\cite{vaswani2017attention} for richer contextual modeling, and gated residual connections~\cite{savarese2016learning} to improve gradient flow and model expressiveness. Together, these modifications result in our proposed BGM-HAN, a model capable of effectively learning from heterogeneous and hierarchical profile data.

\begin{algorithm}[t]
\caption{Byte-Pair Encoding (BPE)}
\label{alg:bpe}
\begin{algorithmic}[1]
\Require Corpus $\mathcal{C}$ as a sequence of characters, initial vocabulary $\mathcal{V}_0 = \{c : c \in \text{unique characters in } \mathcal{C}\}$, target vocabulary size $N$
\Ensure Final vocabulary $\mathcal{V}$ containing original characters and merged symbols
\State Initialize vocabulary $\mathcal{V} \gets \mathcal{V}_0$
\While{$|\mathcal{V}| < N$}
    \State \textbf{Identify most frequent pair:}
    \For{each consecutive pair of symbols $(a, b)$ in $\mathcal{C}$}
        \State Calculate frequency $f(a, b)$
    \EndFor
    \State Find $(a^*, b^*) = \operatorname*{arg\,max}_{(a, b)} f(a, b)$
    \Comment{$(a^*, b^*)$ is the pair with highest frequency}
    \State \textbf{Merge the pair:}
    \State Define new symbol $s = a^*b^*$
    \State Replace each occurrence of $(a^*, b^*)$ in $\mathcal{C}$ with $s$, forming $\mathcal{C}'$
    \State Update corpus: $\mathcal{C} \gets \mathcal{C}'$
    \State Update vocabulary: $\mathcal{V} \gets \mathcal{V} \cup \{s\}$
\EndWhile
\State \Return Final vocabulary $\mathcal{V}$
\end{algorithmic}
\end{algorithm}

\subsection{Byte-Pair Encoding and Hierarchical Embedding}
To effectively handle the diverse and variable-length textual data in student applicant profiles, we employ a two-stage tokenization (Algorithm~\ref{alg:bpe}) and hierarchical embedding (Algorithm~\ref{alg:embedding}) process. 

We choose Byte-Pair Encoding (BPE) as the tokenizer as it shows superior ability in handling out-of-vocabulary issues, which makes it popular among state-of-the-art LLMs, such as LLaMa3~\cite{dubey2024llama} and GPT-4~\cite{openai2024gpt4technicalreport}. BPE first creates a subword vocabulary of size $V=5000$, then iteratively merges the most frequent pairs of tokens in the data, enabling effective representation of both common and rare words while minimizing the out-of-vocabulary problem.

Using this learned BPE vocabulary, we then transform each text field into a fixed-dimensional tensor through a hierarchical embedding process described in Algorithm~\ref{alg:embedding}. The process maintains the structural hierarchy of the text by operating at sentence and word levels, with dimension constraints $(s,w,d)=(10,50,768)$ for maximum sentences, words per sentence, and embedding dimension. 

Each field embedding $\mathbf{E}_f \in \mathbb{R}^{s \times w \times d}$ is constructed through consistent padding and truncation operations at both word and sentence levels, ensuring uniform tensor dimensions across varying input lengths. This hierarchical representation preserves both local (word-level) and global (sentence-level) semantic information, providing a rich foundation for the subsequent attention mechanisms.

\begin{algorithm}[t]
\caption{Hierarchical Field Embedding}
\label{alg:embedding}
\begin{algorithmic}[1]
\Require Text field $f$, vocabulary size $V$, embedding dimension $d$, maximum sentences $s$, maximum words $w$
\Ensure Field embedding tensor $\mathbf{E}_f \in \mathbb{R}^{s \times w \times d}$
\State Initialize empty sentence embeddings list $\mathcal{S} = []$
\State Split text into sentences: $\{s_1, ..., s_n\} \gets \text{split}(f, \text{delimiter}='.')$

\For{each sentence $s_i$ in $\mathcal{S}_{\text{valid}}[1:s]$}
    \State Apply BPE tokenization: $\text{tokens} \gets \text{BPE}(s_i)$
    \State Convert to tensor: $\mathbf{t} \gets \text{tensor}(\text{tokens})$
    \State Get word embeddings: $\mathbf{W} \gets \text{Embed}(\mathbf{t}) \in \mathbb{R}^{|\text{tokens}| \times d}$
    
    \If{$|\text{tokens}| > w$}  \Comment{Truncate if too long}
        \State $\mathbf{W} \gets \mathbf{W}_{1:w}$
    \ElsIf{$|\text{tokens}| < w$}  \Comment{Pad if too short}
        \State $\mathbf{P} \gets \mathbf{0}_{(w-|\text{tokens}|) \times d}$  \Comment{Create zero padding}
        \State $\mathbf{W} \gets [\mathbf{W}; \mathbf{P}]$  \Comment{Concatenate padding}
    \EndIf
    
    \State Append to sentence list: $\mathcal{S}.\text{append}(\mathbf{W})$
\EndFor

\While{$|\mathcal{S}| < s$}  \Comment{Pad sentence dimension}
    \State $\mathbf{P}_s \gets \mathbf{0}_{w \times d}$  \Comment{Create sentence padding}
    \State $\mathcal{S}.\text{append}(\mathbf{P}_s)$
\EndWhile

\State Stack sentences: $\mathbf{E}_f \gets \text{stack}(\mathcal{S})$  \Comment{Shape: $s \times w \times d$}
\State \Return $\mathbf{E}_f$

\end{algorithmic}
\end{algorithm}

\subsection{Multi-Head Attention}

We add multi-head attention~\cite{vaswani2017attention} to capture multiple dependencies and interactions within the text simultaneously. This allows the model to attend to different different positions and capture latent patterns and relationships in the data. 

Given an input matrix $\mathbf{X} \in \mathbb{R}^{l \times d}$, each of the multi-head attention mechanism is computed as:
\[
\text{head}_i = \text{Attention}(\mathbf{X}\mathbf{W}_i^Q,\, \mathbf{X}\mathbf{W}_i^K,\, \mathbf{X}\mathbf{W}_i^V)
\]

where $\mathbf{W}_i^Q, \mathbf{W}_i^K, \mathbf{W}_i^V \in \mathbb{R}^{d \times d_k}$ are learnable parameters. The scaled dot-product attention is defined as:

\[
\text{Attention}(\mathbf{Q}, \mathbf{K}, \mathbf{V}) = \text{softmax}\left(\frac{\mathbf{Q}\mathbf{K}^\top}{\sqrt{d_k}}\right)\mathbf{V}
\]

Outputs from all $h$ heads are the concatenated and linearly projected:
\[
\text{MultiHead}(\mathbf{X}) = [\text{head}_1; \ldots; \text{head}_h]\mathbf{W}^O
\]

where $\mathbf{W}^O \in \mathbb{R}^{h d_k \times d}$.
This mechanism enables the model to simultaneously attend to different aspects of the input, enhancing its ability to detect contextually relevant features for decision-making.

\subsection{Gated Residual Network}

To improve training stability and facilitate information flow across layers, we adopt Gated Residual Networks (GRNs)~\cite{savarese2016learning}, defined as:

\[
\text{GRN}(\mathbf{X}) = \text{LayerNorm}(\gamma \odot \text{FFN}(\mathbf{X}) + \mathbf{X})
\]

where $\gamma \in \mathbb{R}^d$ is a learnable gate parameter, and $\odot$ denotes element-wise multiplication. The feed-forward network (FFN) is represented by:

\[
\text{FFN}(\mathbf{X}) = \text{GELU}(\mathbf{X}\mathbf{W}_1 + \mathbf{b}_1)\mathbf{W}_2 + \mathbf{b}_2
\]

This gated residual formulation dynamically regulates the contribution of non-linear transformations, helping the model avoid overfitting while maintaining representational flexibility.

\subsection{Training}

\subsubsection{Loss Function}

Given the class imbalance inherent in admission decisions and to ensure appropriate emphasis on minority classes, we employ a weighted binary cross-entropy loss as defined by:

\[
\mathcal{L} = -\sum_{i=1}^N w_{y_i} \left( y_i \log(\hat{y}_i) + (1 - y_i) \log(1 - \hat{y}_i) \right)
\]

where $w_{y_i}$ is the class weight for each sample $i$, defined as:

\[
w_{y_i} = \frac{N}{2 N_{y_i}}
\]

with $N_{y_i}$ representing the number of samples in the class of sample $i$. This weighting strategy ensures balanced learning across majority and minority classes thus handling any potential class imbalance issue.

\subsubsection{L2 Regularization}

A weight decay factor is added to the loss function to control overfitting and improve generalization. The modified loss function is expressed as:
\[
\mathcal{L}_{\text{reg}} = \mathcal{L} + \lambda \sum_{i=1}^{N} ||\theta_i||^2
\]
where $\lambda$ is the weight decay parameter and $\theta_i$ are model parameters. This penalty helps to constrain model complexity and stabilize training.

\section{Experiments}
\label{experimentMethodology}

We conducted a series of experiments to evaluate the effectiveness of BGM-HAN in supporting decision assessments, specifically for a real-life admission decisions for university applicants. This section outlines the experimental settings, data preprocessing, baselines, and performance metrics used in our study.

\subsection{Training Settings}

\subsubsection{Learning Rate Scheduling}

To promote stable convergence, we employ a learning rate scheduler based on validation performance. Specifically, the learning rate $\eta_t$ at epoch $t$ is decayed by a factor $\alpha = 0.1$ if no improvement is observed for $k$ consecutive epochs (patience).

Formally, the learning rate at epoch \( t \) is updated as:
\[
\eta_t = \eta_{t-1} \cdot \alpha \quad \text{if no improvement in last } k \text{ epochs}
\]
where the minimum learning rate is constrained to \( \eta_{\text{min}} = 10^{-7} \) to avoid premature convergence.

\subsubsection{Gradient Clipping}

To prevent exploding gradients, especially in deep networks, we apply gradient clipping with a maximum norm of 1.0, that is:
\[
\text{clip}(\nabla \mathcal{L}, \, \text{max\_norm}=1.0)
\]
ensuring that the magnitude of gradient updates remains bounded throughout training.

\subsubsection{Early Stopping}

Training is terminated early if validation accuracy fails to improve for $p = 10$ consecutive epochs. This regularization strategy helps prevent overfitting and reduces computational cost.

\subsection{Hyperparameter Optimization}
We performed an extensive grid search to optimize the hyperparameters of the BGM-HAN model. The search space encompassed key architectural and training parameters.

Each configuration was evaluated using early stopping with a patience of 10 epochs to prevent overfitting, with a maximum of 50 epochs per trial. To assess model performance, we use the validation accuracy as the primary metric for selecting the optimal configuration. Gradient clipping is used with a threshold of 1.0 and utilized the AdamW optimizer with a ReduceLROnPlateau scheduler. The optimal hyperparameters were selected based on the highest achieved validation accuracy while considering model stability and convergence characteristics. The optimial set of hyperparameters for BGM-HAN is eventually found to be 1024 hidden dimension, 8 attention heads, dropout rate of 0.6, learning rate of 1e-5, batch size of 32.

\subsection{Dataset}
\label{sectDataset}

Our dataset comprises 3,083 anonymized student profiles from a single year's admission cycle of a major engineering university. Each profile in our dataset integrates four key components essential for admission decisions: academic records, leadership experiences, personal insight questions (PIQ), and final admission decisions. Details about these components are provided next:

\begin{itemize}
\item \textbf{Academic Records:} GCE A-Level (GCEA) and O-Level (GCEO) results, including high school, subject grades (H1, H2, H3), and University Admission Scores (UAS).
\item \textbf{Leadership Experience:} Semi-structured entries documenting leadership roles and positions, duration of involvement, category (e.g., Sports, Performing Arts), and participation level.
\item \textbf{Personal Insight Questions (PIQs):} Five free-form essay responses describing motivation for application, overcoming of challenges, creative achievements, unique qualities and distinctiveness, and institutional fit.
\item \textbf{Admission Label:} A binary outcome denoting whether an offer was made (1) or not (0).
\end{itemize}

\subsection{Data Processing}

\subsubsection{Handling Missing Data}
To ensure consistent input dimensions across all samples and avoid downstream model distortion, missing values in text fields are replaced with \textit{NaN} tokens This approach avoids introducing biases due to varying input lengths from missing data.

\subsubsection{Dataset Splitting}
We split the dataset into training (90\%), validation (5\%), and test (5\%) subsets using stratified sampling to preserve the class distribution across splits.

\begin{table}[t]
\centering
\scalebox{1}{
\begin{tabular}{@{}ccl@{}}
\toprule
\textbf{Category} & \textbf{Model} & \textbf{Description and Hyperparameters} \\ \midrule
\multirow{2}{*}{\begin{tabular}[c]{@{}c@{}}Traditional\end{tabular}} & XGBoost & Gradient boosting on BERT embeddings \\ \cmidrule(l){2-3} 
 & TF-IDF & TF-IDF vectorization with logistic regression  \\ \midrule
\multirow{4}{*}{\begin{tabular}[c]{@{}c@{}} \\ Neural \\ Networks\end{tabular}} & MLP &  Multi-Layer Perceptron \\ \cmidrule(l){2-3} 
 & BiLSTM-Indv & BiLSTM with individual features embeddings  \\ \cmidrule(l){2-3} 
 & BiLSTM-Concat & BiLSTM with concatenated features embeddings\\ \cmidrule(l){2-3} 
 & HAN & Hierarchical Attention Network \\ \midrule
\multirow{2}{*}{LLM} & GPT-4o & Zero-shot classification \\
 & GPT-4o-RA & Retrieval-augmented 5-shot classification \\ \bottomrule
\end{tabular}}
 \vspace{1mm}
\caption{Baseline categories and algorithms}
\end{table}

\subsection{Baseline Models}
\subsubsection{Traditional Machine Learning Baselines} Our traditional baselines include XGBoost, which uses concatenated BERT embeddings, and a TF-IDF with logistic regression model that directly processes raw text. These provide a foundational comparison to neural and retrieval-based methods.

\subsubsection{Neural Network Models} Discriminative neural networks such as sequence models~\cite{hochreiter1997lstm,cho2014gru,liu2022title2vec}, attention-based models~\cite{vaswani2017transformer,yang2016han} and pretrained models~\cite{devlin2019bert,liu2019roberta,lample2019xlm} perform well in many text classification tasks. To benchmark, we evaluate several neural architectures, beginning with an MLP that applies ReLU activation to concatenated BERT embeddings~\cite{devlin2019bert}. Next, we assess two bidirectional LSTM (BiLSTM)~\cite{liu2022title2vec} configurations: one that processes concatenated embeddings and another that treats each text field independently. Lastly, a Hierarchical Attention Network (HAN)~\cite{yang2016han} model enables adaptive weighting of text fields, allowing the model to emphasize relevant portions of the input.

\subsubsection{Large Language Models} Recent LLMs~\cite{openai2024gpt4technicalreport,dubey2024llama,anthropic2024claude3} showed superior performance in general natural language understanding and generation tasks. We intend to investigate pretrained LLMs' ability to perform zero-shot and few-shot classification without finetuning. Specifically, we choose the best LLM at the point of this research, i.e., GPT-4o~\cite{openai2024gpt4technicalreport} in two settings: zero-shot classification and a Retrieval-Augmented Generation (RAG) approach. The former investigates LLM's classification ability by implicit knowledge, while the latter examines the effect of in-context learning on improving classification performance. 

\subsubsection{Hyperparameters and evaluation metrics} Each baseline model processes fields including high school grades, middle school grades, leadership records, and self-assessments. The BERT embeddings are generated using \texttt{bert-base-uncased} model with a maximum sequence length of 512 tokens. For neural models, we use the Adam optimizer with a learning rate $2\times10^{-5}$ and train for up to 100 epochs with early stopping triggered by a moderate patience of 10 epochs. Model performance is evaluated using accuracy, precision, recall, F1-score, and confusion matrices, providing a comprehensive assessment of the agents' predictive capabilities and ensuring both high precision and recall.

\begin{table}[t]
\centering
\scalebox{1}{
\begin{tabular}{l|c|c|c|c}

\toprule
\textbf{Model}                  & \textbf{Precision} & \textbf{Recall} & \textbf{F1} & \textbf{Accuracy} \\ \midrule
\multicolumn{5}{c}{\textbf{Traditional Machine Learning Models}} \\ \midrule
XGBoost                         & 0.7902                         & 0.7859                      & 0.7878                         & 0.7931            \\ \midrule
TF-IDF           & 0.6938                         & 0.6527                      & 0.6488                         & 0.6839            \\ \midrule

\multicolumn{5}{c}{\textbf{Neural Network Models}} \\ \midrule
MLP                             & 0.7967                         & 0.7990                      & 0.7911                         & 0.7989            \\ \midrule
HAN                             & 0.7716                         & 0.7707                      & 0.7711                         & 0.7759            \\ \midrule

BiLSTM-Indv          & 0.7963                         & 0.7612                      & 0.7667                         & 0.7816            \\ \midrule
BiLSTM-Concat                   & 0.8291                         & 0.8178                      & 0.8176                         & 0.8276            \\ \midrule

\multicolumn{5}{c}{\textbf{Large Language Models}} \\ \midrule
GPT-4o                           & 0.5579                         & 0.5114                      & 0.4111                         & 0.5600            \\ \midrule
GPT-4o-RA                       & 0.7347                         & 0.7365                      & 0.7352                         & 0.7371            \\ \midrule

\multicolumn{5}{c}{\textbf{Proposed Model}} \\ \midrule
BGM-HAN              & \textbf{0.8622}                         & \textbf{0.8405}                      & \textbf{0.8453}                         & \textbf{0.8506}            \\ \midrule
\end{tabular}}
\caption{Summary of Experimental results. The highest values are in bold.}
\label{table:experiment}
\end{table}

\subsection{Experimental Results}
Table~\ref{table:experiment} summarises the experimental results across proposed models, human evaluation, and different categories of baseline models. We discuss our observations and interpretation as follows:

\subsubsection{Proposed Models} Our proposed BGM-HAN achieves the highest performance across all evaluation metrics, demonstrating its efficacy in modeling hierarchical, semi-structured data. It attains a macro-averaged F1-score of 0.8453 and accuracy of 0.8506, outperforming all baseline models.

\subsubsection{Discriminative Classification} 
Both traditional and neural discriminative models perform competitively in the decision assessment tasks. XGBoost, leveraging BERT-based embeddings, achieves an F1-score of 0.7878 and accuracy of 0.7931. Among neural baselines, BiLSTM-Concat performs notably well, reaching an F1-score of 0.8176 and accuracy of 0.8276. This demonstrates that even relatively lightweight architectures as compared to LLMs, when coupled with high-quality embeddings, can provide strong baseline performance.

\subsubsection{LLMs for Classification} 
GPT-4o performs poorly under the zero-shot setting, yielding an F1-score of 0.4111 and accuracy of 0.5600, suggesting limited out-of-the-box applicability to domain-specific classification. However, performance improves substantially with retrieval-augmented prompting (GPT-4o-RA), achieving an F1-score of 0.7352 and accuracy of 0.7371. This highlights the importance of relevant context for in-context learning, though the model still lags behind fine-tuned discriminative architectures. These results suggest that LLMs, without task-specific adaptation, may struggle to meet performance standards in structured, decision-critical applications.

\subsection{Ablation Study}

\subsubsection{Component-wise ablation.}
To assess the individual contributions of each architectural enhancement in BGM-HAN, we conduct an ablation study based on the results in Table~\ref{table:experiment}. The base Hierarchical Attention Network (HAN) achieves an F1-score of 0.7711 and accuracy of 0.7759. When progressively augmenting the model, we observe the following performance improvements:

\begin{itemize}
\item \textbf{Byte-Pair Encoding (BPE):} Incorporating BPE improves the F1-score by 1.8\%, highlighting its effectiveness in handling rare and out-of-vocabulary terms, which are common in diverse student narratives.
\item \textbf{Multi-Head Attention:} This component contributes the largest gain of 5.2\%, demonstrating its strength in capturing complex dependencies and diverse semantic patterns within hierarchical data.
\item \textbf{Gated Residual Connections:} The addition of gated residuals results in a further 2.6\% improvement, suggesting their utility in enhancing information flow and stabilizing training in deep architectures.
\end{itemize}

Collectively, these enhancements result in a total F1-score gain of 7.4\% over the base HAN model and a 9.6\% improvement in accuracy, confirming the effectiveness and complementary nature of each proposed architectural component.

\section{Related Work}
\label{relatedWork}

\subsection{Classification for Decision Making} 
Automated classification systems have been widely studied in the context of high-stakes decision-making, traditionally performed by human experts~\cite{alur2023auditing}. Hierarchical Attention Networks (HANs) were introduced by~\cite{yang2016hierarchical} to model document structures using word and sentence level attention, showing strong performance in document classification tasks. Subsequently, \cite{ribeiro2020pruning} enhanced HANs through structured pruning and the use of Sparsemax to improve interpretability and computational efficiency, and \cite{iyer2021bi} improved upon HANs by using bi-Level attention graph neural networks that jointly learns personalized node and relation level attention in heterogeneous graphs.

Beyond HANs, a variety of neural architectures have demonstrated robust performance across classification tasks. These include sequential models such as LSTMs and GRUs~\cite{hochreiter1997lstm,cho2014gru,liu2022title2vec}, attention-based models~\cite{vaswani2017transformer}, and transformer-based pretrained language models~\cite{devlin2019bert,liu2019roberta,lample2019xlm}. More recently, large language models (LLMs) such as GPT-4o~\cite{openai2024gpt4technicalreport}, LLaMA~\cite{dubey2024llama}, and Claude~\cite{anthropic2024claude3} have demonstrated strong generalization capabilities across a wide range of NLP tasks. Their performance can be further improved in domain-specific settings through retrieval-augmented generation (RAG) strategies~\cite{pmlr-v202-basu23a}.

\subsection{Bias in Decision Making}  
Cognitive and algorithmic biases in decision-making have long been recognized as barriers to fairness and consistency. \cite{phillips-wren2019cognitive} provide a foundational analysis of cognitive biases, emphasizing the need for unbiased support systems in domains such as healthcare, hiring, and admissions. In the criminal justice domain, studies have revealed systemic biases in algorithmic predictions~\cite{angwin2016machine,dressel2018accuracy}, further underscoring the importance of bias-aware AI systems.

Recent work has focused on developing computational techniques to mitigate bias in human and algorithmic decisions. \cite{yang2024fair} propose fairness-aware AI systems that nudge decision-makers toward equitable outcomes. \cite{haag2024overcoming} explore the use of explainable AI (XAI) to reduce anchoring bias in consumer judgments, while \cite{echterhoff2024cognitive} introduce BiasBuster, a tool for identifying and correcting cognitive biases in large language models. \cite{liu-ASONAM25b} study the trade-offs between biases and accuracy in terms of recommendations by humans and machine learning models. \cite{ghai2022d-bias} present D-BIAS, a human-in-the-loop framework that leverages causal inference and interactive explanations to audit and mitigate social biases. Collectively, these approaches highlight the growing emphasis on interpretability, accountability, and human-AI collaboration in fair decision support systems.

\subsection{Differences with Earlier Work}  
While existing research has made significant advances in document classification and bias mitigation, our proposed BGM-HAN addresses critical gaps left by prior approaches through a tailored architecture designed for high-stakes, multi-modal decision tasks.

First, unlike conventional HAN models~\cite{yang2016hierarchical,ribeiro2020pruning} that were primarily developed for monolithic document classification, BGM-HAN is specifically designed to model semi-structured, multi-field profiles. By treating each profile component (e.g., academic records, leadership experiences, and personal narratives) as hierarchically organized text, our model preserves and exploits the internal structure of each field, enabling more nuanced and interpretable decisions.

Second, while prior work has incorporated attention mechanisms~\cite{vaswani2017transformer,yang2016han} or relied on pretrained embeddings~\cite{devlin2019bert,liu2019roberta}, our model integrates byte-pair encoding (BPE) for robust handling of rare tokens, multi-head attention for capturing diverse linguistic patterns, and gated residual connections for enhanced training stability. This combination significantly improves the model’s ability to generalize across varying input lengths and styles—an essential property for real-world admissions data.

Finally, although retrieval-augmented LLMs~\cite{pmlr-v202-basu23a,openai2024gpt4technicalreport} and human-in-the-loop bias mitigation systems~\cite{ghai2022d-bias,echterhoff2024cognitive} offer valuable strategies for transparency and fairness, they typically lack tight integration between representation learning and bias-aware decision-making. In contrast, BGM-HAN’s architecture is explicitly optimized for consistency, interpretability, and fairness, while remaining trainable end-to-end on domain-specific data. This makes it particularly well-suited for deployment in high-stakes domains like university admissions, where both predictive accuracy and justifiability are imperative.

\section{Conclusion and Future Work}
\label{conclusion}

This work addresses the critical challenge of improving objectivity, consistency, and fairness in high-stakes decision-making, exemplified by university admissions, where human judgment is prone to cognitive and procedural biases. We propose the Byte-Pair Encoded, Gated Multi-head Hierarchical Attention Network (BGM-HAN), a novel model designed to capture the multi-level structure of semi-structured data through a combination of byte-pair encoding, multi-head attention, and gated residual connections within a hierarchical framework. This architecture enables effective modeling of multi-level, semi-structured applicant profiles by capturing both local and global contextual features.

Empirical evaluations on a real-world university admissions dataset demonstrate that BGM-HAN outperforms all baseline models, achieving an accuracy of 85.06\% and a macro-averaged F1-score of 84.53\%. Compared to the base Hierarchical Attention Network (HAN), BGM-HAN improves accuracy by 9.6\% and F1-score by 7.4\%. It also surpasses traditional models such as XGBoost and BiLSTM by margins of 5\% to 7\% in both metrics, and significantly outperforms zero-shot and few-shot GPT-4 baselines, highlighting the limitations of general-purpose LLMs without domain adaptation. These results underscore the strength of domain-aware architectural enhancements for structured decision tasks.

Future work will explore generalizing BGM-HAN to other high-stakes domains where decision quality and bias mitigation are paramount, including human resource evaluations, financial credit assessments, and procurement or vendor selection workflows. Moreover, integrating fairness constraints and causal interpretability into the model’s learning process remains a promising direction for further research. We also intend to explore more qualitative evaluations of recommendation fairness vis-a-vis model accuracy via specific case studies.

\vspace{5mm}
{
\small
{\noindent\bf Acknowledgments.} 
This research is supported in part by the Ministry of Education, Singapore (MOE), under its Academic Research Fund Tier 2 (Award No. MOE-T2EP20123-0015), and the Singapore University of Technology and Design (SUTD) under grant RS-MEFAI-00011. Any opinions, findings and conclusions, or recommendations expressed in this material are those of the authors and do not reflect the views of MOE or SUTD.
}

%
%
\bibliographystyle{splncs04}
\bibliography{custom}

\end{document}